%% file: CITDS_paper.tex
\documentclass[a4paper,conference]{IEEEtran}
\IEEEoverridecommandlockouts
\usepackage{cite}
\usepackage{amsmath,amssymb,amsfonts}
\usepackage{algorithmic}
\usepackage{graphicx}
\usepackage{textcomp}
\usepackage{xcolor}
\usepackage{comment}

\usepackage{mathtools}
\usepackage{multirow}
\usepackage{xspace}
\usepackage{makecell}
\usepackage{array}
\usepackage{booktabs}

\def\BibTeX{{\rm B\kern-.05em{\sc i\kern-.025em b}\kern-.08em
    T\kern-.1667em\lower.7ex\hbox{E}\kern-.125emX}}
    
\usepackage[caption=false, font=footnotesize]{subfig}
\input{00_macros.tex}

\IEEEoverridecommandlockouts\IEEEpubid{\makebox[\columnwidth]{978-1-6654-
9653-7/22/\$31.00~\copyright~2022 IEEE \hfill}
\hspace{\columnsep}\makebox[\columnwidth]{ }}

\begin{document}
\title{Negative Sampling in Variational Autoencoders 
\thanks{This work was supported by the Hungarian Ministry of Innovation and Technology NRDI Office within the framework of the Artificial Intelligence National Laboratory Program and by the project TKP2021-NKTA-62.}}

\author{
    \IEEEauthorblockN{
        Adri\'an Csisz\'arik \IEEEauthorrefmark{1} \IEEEauthorrefmark{2},
        Beatrix Benkő \IEEEauthorrefmark{1} \IEEEauthorrefmark{2},
        D\'aniel Varga \IEEEauthorrefmark{1}}
    \and\and
    \IEEEauthorblockA{
        \centering
        \hspace{2.5cm}
        \begin{tabular}{c c}
            \begin{tabular}{@{}c@{}}
                \IEEEauthorrefmark{1}
                \textit{Alfr\'ed R\'enyi Institute of Mathematics}\\
                Budapest, Reáltanoda u. 13-15., Hungary \\
            \end{tabular} & 
            \begin{tabular}{@{}c@{}}
                \IEEEauthorrefmark{2}
                \textit{Eötvös Loránd University}\\
                Budapest, Pázmány Péter s. 1/C, Hungary
            \end{tabular}
        \end{tabular}
    }
    \hspace{2cm}\{csadrian, bbea, daniel\}@renyi.hu
}

\maketitle

\begin{abstract}
Modern deep artificial neural networks have achieved great success in the domain of computer vision and beyond. However, their application to many real-world tasks is undermined by certain limitations, such as overconfident uncertainty estimates on out-of-distribution data or performance deterioration under data distribution shifts. 
Several types of deep learning models used for density estimation through probabilistic generative modeling have been shown to fail to detect out-of-distribution samples by assigning higher likelihoods to anomalous data. 
We investigate this failure mode in Variational Autoencoder models, which are also prone to this, and improve upon the out-of-distribution generalization performance of the model by employing an alternative training scheme utilizing negative samples. We present a fully unsupervised version: when the model is trained in an adversarial manner, the generator's own outputs can be used as negative samples.
We demonstrate empirically the effectiveness of the approach in reducing the overconfident likelihood estimates of out-of-distribution inputs on image data.
\end{abstract}


\input{1_introduction}

\input{2_background}

\input{3_training_with_ns}

\input{4_experimnetal_results}

\input{5_related_work}

\input{6_conclusion}

\bibliographystyle{IEEEtran}
\bibliography{IEEEabrv,CITDS_references}

\end{document}

%% file: 00_macros.tex
\usepackage{amsmath,amsfonts,bm}

\def\eqref#1{equation~\ref{#1}}

\def\1{\bm{1}}

\def\vx{{\bm{x}}}

\def\vz{{\bm{z}}}

\DeclareMathAlphabet{\mathsfit}{\encodingdefault}{\sfdefault}{m}{sl}
\SetMathAlphabet{\mathsfit}{bold}{\encodingdefault}{\sfdefault}{bx}{n}

\newcommand{\E}{\mathbb{E}}

\newcommand{\KL}{D_{\mathrm{KL}}}

\renewcommand{\vec}[1]{\mathbf{#1}}

\newcommand{\vxi}[0]{\vec{x}^{(i)}}
\newcommand{\vxineg}[0]{\vec{\overline{x}}^{(i)}}
\newcommand{\vxigen}[0]{\vec{\hat{x}}^{(i)}}

\newcommand{\ptzxi}[0]{p_{\theta}(\vz|\vxi)}
\newcommand{\ptzx}[0]{p_{\theta}(\vz|\vx)}

\newcommand{\approxposteriornegi}[0]{q_{\phi}(\vz|\vxineg)}

\newcommand{\jointlikelihoodnegi}[0]{p_{\theta}(\vxineg, \vz)}

\newcommand{\ptxiz}[0]{p_{\theta}(\vxi|\vz)}

\newcommand{\ptxhatiz}[0]{p_{\theta}(\vec{\hat{x}}^{(i)}|\vz)}

\newcommand{\ptxz}[0]{p_{\theta}(\vx|\vz)}

\newcommand{\qpzxi}[0]{q_{\phi}(\vz|\vxi)}
\newcommand{\qpzx}[0]{q_{\phi}(\vz|\vx)}

\newcommand{\qpzxnegi}[0]{q_{\phi}(\vz|\vxineg)}
\newcommand{\qpzxgeni}[0]{q_{\phi}(\vz|\vxigen)}

\newcommand{\ptx}[0]{p_{\theta}(\vx)}
\newcommand{\ptxi}[0]{p_{\theta}(\vxi)}
\newcommand{\pzbar}[0]{\bar{p}(\vz)}
\newcommand{\pvzbar}[0]{\bar{p}(\vz)}

\newcommand{\pt}[0]{p_{\theta}}
\newcommand{\qp}[0]{q_{\phi}}

\DeclarePairedDelimiterX{\infdivx}[2]{(}{)}{#1\;\delimsize\|\;#2}

\newcommand{\kl}[2]{\ensuremath{\KL \infdivx{#1}{#2}}\xspace}
\DeclareMathOperator{\ELBO}{ELBO}
\DeclareMathOperator{\EUBO}{EUBO}

%% file: 1_introduction.tex
\section{Introduction}

Learning semantically meaningful and useful representations of high dimensional data --- such as natural images --- for downstream tasks in an unsupervised manner is a big promise of probabilistic generative modeling. While a plethora of work demonstrates the effectiveness of deep generative models in this regard, recent work of \cite{nalisnick2018deep, choi2018waic} show that these models often fail even at a task that is supposed to be close to their original goal of learning densities. As highlighted by \cite{nalisnick2018deep}, current deep generative models like Variational Autoencoders \cite{diederik2014auto}, PixelCNNs \cite{van2016pixel} and flow-based models \cite{kingma2018glow} consistently assign higher likelihoods to elements from an unseen image dataset supposedly belonging to different distribution than to the samples from the learned data distribution itself. Therefore, these models cannot be used to distinguish inlier and outlier samples simply by setting a threshold on the likelihood. As generative models are becoming more and more ubiquitous due to the massive progress in this area, 
in order to advance their applicability, it is of fundamental importance to understand these phenomena.

In this work we study Variational Autoencoder (VAE) models \cite{diederik2014auto}, and besides the likelihood estimates, we also investigate to what extent the latent representation of a data point can be used to identify out-of-distribution (OOD) samples  --- data points that do not belong to the true data distribution that we aim to model. For this purpose, we consider the Kullback-Leibler (KL) divergence between the prior and the posterior distribution of a data point as a score to distinguish inliers and outliers.
Our contributions are summarized as follows:
\begin{itemize}
    \item We demonstrate empirically that the extent of the notorious phenomenon of counter-intuitively high out-of-distribution likelihood estimates present in VAEs largely depends on the observation model of the VAE. In particular, our experiments show that with certain datasets it diminishes  when a Gaussian noise model is considered 
    instead of a Bernoulli. 
    \item We propose \textit{negative sampling in Variational Autoencoders} as an approach to alleviate the above weaknesses of the model family. Our modified training procedure seeks to minimize the likelihood of negative samples, while still maximizing the likelihood of training samples. 
    Negative samples can be obtained from an auxiliary dataset, or --- to remain completely in the unsupervised setting --- from the generative model itself trained adversarially on the ground truth data distribution.
    \item We present empirical evidence that using negative samples either from an auxiliary dataset or from an adversarial training scheme     significantly improves the discriminative power of VAE models regarding OOD samples.
\end{itemize}

The general intuition behind our training approach is that if the posterior distribution of each and every point is pulled towards the prior in the VAE model, then it is rather natural to expect that the system will map OOD samples close to the prior, as well. This viewpoint suggests that providing negative signals throughout the learning process would be beneficial to enhance the OOD discriminative power of the system. 

Reference \cite{hendrycks2018deep} demonstrates that using auxiliary datasets as source of OOD examples for supervisory signal significantly improves the performance of existing anomaly detection models on image and text data. First, we study how this approach can be employed in the VAE setting. Beyond that, we propose a method which remains completely in the unsupervised learning paradigm.
The crux of this approach is to use a generative model to provide near-manifold negative samples throughout the training process for which the model is either implicitly or explicitly encouraged to give low likelihood estimates. 

%% file: 2_background.tex
\section{Background}

The generative modeling task aims to model a ground truth data density $p^{*}(\vx)$ on a space $\mathcal{X}$ by learning to generate samples from the corresponding distribution. The learning is done in an unsupervised manner with sampled observables $\mathbf{X} = \{\vxi\}_{i=1}^{N} \subseteq \mathcal{X}$ as training points assumed to be drawn independently from $p^{*}(\vx)$, where $N \in \mathbb{N}$ is the sample size and $\mathcal{X}$ denotes the sample space. In latent variable models, the observables are modeled together with hidden variables $\vz$ on which a prior distribution $p(\vz)$ is imposed.

The Variational Autoencoder (VAE) \cite{diederik2014auto} is a latent variable generative model that takes the maximum likelihood approach and maximizes a lower bound of the sample data log likelihood $\sum_{i=1}^{N}$ $\log p_{\theta}(\vxi)$, where $\theta$ are the generator parameters. The utilized Evidence Lower Bound (ELBO) $\mathcal{L}(\theta,\phi,\vxi)$ comes from a variational approximation $\qpzxi$ of the intractable posterior $\ptzxi$, where $\phi$ are the variational parameters:
\begin{gather*}
\log\ptxi = \log\int p_{\theta}(\vxi|\vz) p(\vz) \geq \\
\geq \underbrace{\E_{\qpzxi} \log\ptxiz}_{\text{reconstruction term}} - \underbrace{\kl{\qpzxi}{p(\vz)}}_{\text{KL divergence term}} \triangleq \\ \triangleq \mathcal{L}(\theta,\phi,\vxi).
\end{gather*}
In the VAE setting, the parametrized distributions $\pt$ and $\qp$ are modeled with neural networks and are trained jointly to maximize $\mathcal{L}$ on training data with some variant of the Stochastic Gradient Descent.

The prior $ p(\vz)$ is often chosen to be the multivariate standard normal distribution, and a Bernoulli or Gaussian noise model is considered in the observable space to define the likelihood. To give likelihood estimates for unseen data points at test time, one can use the trained inference model $\qpzx$ (also referred to as encoder) and generative model $\ptxz$ (also referred to as decoder) to estimate the ELBO, thus giving a lower bound of the log-likelihood.

%% file: 3_training_with_ns.tex
\section{Training with negative samples}

Let denote $\mathbf{\overline{X}} = \{\vxineg\}_{i=1}^{M} \subseteq \mathcal{X}$ the negative samples with sample size $M \in \mathbb{N}$.
One can acquire such negative samples from various sources, as discussed later in this section. First, we consider incorporating them into the VAE model training in the context of the maximum likelihood framework.
With this intention, we arrive at the idea of minimizing the likelihood with the
negative samples, while following the maximum likelihood principle with the positive samples.

\subsection{Using a negative prior}
\label{subsection:negprior}

Perhaps the simplest way to incorporate negative samples in the VAE training process is by introducing an additional prior distribution $\pzbar$ (called the \emph{negative prior}) on the latent variables $\vz$ into which the representations of \emph{negative samples} are meant to be mapped by the inference model. This can be encouraged during the training process by adding to the regular ELBO a new loss term: the KL divergence of the posterior distributions of negative samples to this negative prior. Thus the joint loss function to be minimized  is as follows:
\begin{alignat*}{2}
\mathcal{L}_{np} (& \theta,\phi,\vxi,\vxineg) \triangleq \\ \triangleq &  -\mathcal{L}(\theta,\phi,\vxi)+ \kl{\qpzxnegi}{\pvzbar} = \\ = & \underbrace{-\E_{\qpzxi} \log\ptxiz+\kl{\qpzxi}{p(\vz)}}_{-1 \cdot \text{ ELBO for }\vxi} \\ & + \underbrace{\kl{\qpzxnegi}{\pvzbar}}_{\text{KL term for negative }\;\vxineg}.
\end{alignat*}

The loss function $\mathcal{L}_{np}$ is still an upper bound of the negative data log likelihood (for the positive samples) as the added loss term is non-negative. The new loss term explicitly imposes the discriminative task for the inference model: to distinguish inliers and outliers in the latent space. With these two components, while still preserving the aim of maximizing the likelihood for inliers, we also expect the implicit behavior of reducing likelihood estimates for outliers  when the optimization task is 
$ \displaystyle \min_{\theta, \phi} \, - \mathcal{L}_{np}(\theta,\phi,\vxi,\vxineg).$
For the outliers, a trained inference model produces latent representations that are close to the negative prior $\bar{p}(\vz)$, thus, supposedly far from the prior $p(\vz)$. Also, the system is not encouraged to learn to generate from the vicinity of the negative prior, therefore not only the KL term of the likelihood, but the reconstruction part of a negative sample is affected when inferring the likelihood estimate of an outlier.

Importantly, one has numerous options to choose the positive and negative priors. We simply choose to use a standard normal for the positive prior, and a shifted standard normal for the negative prior. 
Worth to note that the magnitude of KL divergence between the negative and the positive prior plays an important role. Larger $\kl{\bar{p}(\vz)}{p(\vz)}$ values result in larger $\kl{\qpzxnegi}{p(\vz)}$ terms when evaluating the KL divergence term of the likelihood in a trained model, and also result in heavier weighted KL divergence terms during the optimization process. E.g., with a farther shifted negative prior mean, a larger penalty is given for a wrong inference.

\subsection{Combining evidence lower and upper bounds}

The VAE model utilizes the ELBO to maximize the likelihood of training data. In an analogous manner, one can use an \emph{upper bound} of evidence $\ptx$ to \emph{explicitly minimize} the likelihood for negative samples, as described in \cite{daniel2019deep}.
While in \cite{daniel2019deep} a $\chi$-divergence based upper bound is used, we experiment with using the Evidence Upper Bound (EUBO) introduced in \cite{EUBO}.
Reference \cite{EUBO} describes the EUBO in the general context of variational inference with the idea of sandwiching the model evidence with lower and upper bound for better model selection. We write up both the ELBO and the EUBO for a data point $\vx \in \mathcal{X}$:
\begin{alignat*}{3}
    & \ELBO = \E_{\qpzx} \left[ \log \left( \frac{\pt(\vx,\vz)}{\qpzx} \right) \right] \leq \log\ptx
    \\ & \log\ptx \leq  
    \E_{\ptzx} \left[ \log \left( \frac{\pt(\vx,\vz)}{\qpzx} \right) \right] = \EUBO.
\end{alignat*}

The EUBO has favourable properties: has mass covering effect advantageous in the approximation of the posterior, and it provides a tighter bound than the upper bounds introduced by Rényi's $\alpha$-divergence or the $\chi$-divergence \cite{EUBO}. However, the EUBO depends on the the true posterior that is intractable in general, so as we start to turn our attention to practicalities, we change the integration with respect to $\qpzx$, and use the Bayes rule to replace the true posterior. The upper bound loss used with negative samples in our experiments is:
\begin{alignat*}{2}
    \overline{\mathcal{L}}(\theta,\phi,\vxineg)  & \triangleq \E_{\qpzxnegi} 
    \left[
    \frac{\jointlikelihoodnegi }{\approxposteriornegi}
    \log \frac{\jointlikelihoodnegi }{\approxposteriornegi} \right].
\end{alignat*}
Combining the lower and upper bound also results in an objective which preserves the aim of maximizing the likelihood for inliers, while also reducing likelihood estimates for outliers. With this intent we use an amortized training and inference scheme with the following joint optimization task:
$\displaystyle \min_{\theta, \phi} \, - \mathcal{L}(\theta,\phi,\vxi) + \overline{\mathcal{L}}(\theta,\phi,\vxineg)$.

We achieved similarly improved OOD results with both the usage of a negative prior and the usage of combined lower and upper bounds, in Section~\ref{section:experiments} we only report the results obtained with the first. The success of both methods suggests that harnessing negative samples and imposing the discriminative task in the latent space to differentiate between inliers and outliers results in better OOD generalization in general.

\subsection{Sources of negative samples}

The choice of the negative samples is an important modeling decision, as it greatly influences the representations learnt by the model. One can come up with many ideas to produce negative samples, for example these can be:
\begin{itemize}
    \itemsep0em
    \item samples from an auxiliary dataset;
    \item the ground truth data with noise added (e.g., Gaussian);
    \item generated samples from the trained model itself;
    \item hard negative samples specifically produced for this purpose by another model.
\end{itemize}
Regarding out-of-distribution detection, the ultimate purpose of the training scheme is to use the negative samples to generalize as much as possible to detect OOD samples by pushing down the likelihood estimates of those. Depending on the source of negative samples, this generalization can be easier or harder. Negative samples that are very far from the data manifold do not facilitate generalization. Noise added to data points is a simple and principled way to sample from the vicinity of the data manifold, but in our setup and experiments we present in Section~\ref{section:experiments}, it does not allow for good generalization. We argue that the reason for this is that discriminating between noisy and noiseless points is too easy for the encoder, so ``semantically'' the noisy versions are far from the data manifold. In contrast, using samples from an auxiliary dataset or samples produced by a generative model (which could be the trained generative model itself) is a more suitable way to acquire near-manifold negative samples, as we will experimentally demonstrate.

\subsection{Adversarial training with generated negatives}

When experimenting with generated samples as negatives, we use an adversarial training scheme where the generator --- and only the generator --- gets an additional gradient signal through the encoder to map the randomly generated images into the prior. This is encouraged via the following additional loss term:
$$ \mathcal{L}_{adv}(\phi,\vxigen) = \kl{\qpzxgeni}{p(\vz)},$$
where $\vxigen$ denotes a generated image obtained from the generator $\ptxhatiz$, and $\vz$ is sampled from the prior $p(\vz)$. Together with the fact that the encoder also gets the generated images as negative samples, this results in an adversarial training procedure, where the discrimination happens in the latent space with the KL divergence.

%% file: 4_experimnetal_results.tex
\section{Experimental results}
\label{section:experiments}

Our main concern is the discriminative power of VAE models regarding OOD samples. Following the conventions of related work, the general experimental setup is as follows: we train a model on a train set of a dataset (e.g. train set of Fashion-MNIST) and then require the model to discriminate between the test set of the train dataset (e.g. test set of Fashion-MNIST) and the test set of an out-of-distribution dataset (e.g. test set of MNIST). During the training phase, the models do not encounter examples from the OOD dataset. Only at test time are they expected to able to distinguish between inliers and OOD samples. This setup mimics the real-word scenario where one has access to a single unlabeled dataset, and has to identify outlier samples. Baseline VAE models fail at this task even in this fabricated setting where there is a clear difference between inliers and outliers, hence our focus on verification in this setup.   

For quantitative assessment, we use the threshold independent Area Under ROC-Curve (AUC) metric calculated with the bits per dimension (BPD) score (denoted by AUC BPD) and also with the KL divergence of the posterior distribution of a data point to the prior (denoted by AUC KL). The employed BPD score defined as 
$\displaystyle BPD(\vx) = \frac{-\log(\ptx)}{dim(\vx) \cdot \log(2)}$ 
is directly calculated from the ELBO. 
All reported numbers in this section are averages of 10 independent runs with standard deviations denoted in parentheses. 

\input{table_grayscale_summary}

\subsection{Experimental setup details}

We conduct experiments on two sets of datasets: color images of size 32x32 (CIFAR-10 \cite{Krizhevsky09learningmultiple}, SVHN \cite{netzer2011reading}, downscaled ImageNet \cite{van2016pixel}) and grayscale images of size 28x28 (MNIST \cite{lecun2010mnist}, Fashion-MNIST \cite{xiao2017fashion}, EMNIST-Letters \cite{cohen_afshar_tapson_schaik_2017}). 
We apply no preprocessing step other than normalizing the input images to $[0,1]$. When an auxiliary dataset is used, in case of grayscale images it is the one out of the three datasets that is not used neither as inlier nor for OOD testing purposes. In case of color images, the auxiliary dataset is the downscaled ImageNet in the reported experiments.

Following \cite{nalisnick2018deep}, for grayscale images, we use the encoder architecture described in \cite{rosca2018distribution}. Also, as in \cite{rosca2018distribution}, all of the models are trained with the RMSProp optimizer with learning rate set to $10^{-4}$. We train the models for 100 epochs with a mini-batch size of $50$. We update the parameters of the encoder and decoder network in an alternating fashion.

For color images we use a DCGAN-style CNN architecture with Conv--Batch\-Norm--ReLU modules for both the encoder and the decoder. The size of the kernels are $4 \times 4$, and the number of filters are $32, 64, 128, 256$ for the encoder; and $128, 64, 32$ for the decoder. In the encoder, the result of the last convolutional layer is flattened, and then two dense layers produce the parameters of the posterior distribution. 
In the decoder, from the latent vector a dense layer produces a $4096$ dimensional vector which is then reshaped to a $256 \times 4 \times 4$ sized tensor before the convolutions, and at the top, a convolution with $3$ filters for the three RGB color channels produces the output. 
All of the models are trained with the Adam optimizer ($\beta_1=0.9, \beta_2=0.999$) for 100 epochs with mini-batch size $50$. The learning rate is set to $10^{-4}$. Again, we update the parameters of the encoder and decoder network in an alternating fashion. 
When generated images are used as negative samples, we employ  spectral normalization \cite{Miyato2018SpectralNorm} for the convolutional weights of the encoder in order to stabilize and enhance the performance of the respective models, and in this case the models are trained for 300 epochs. 
When using negative samples, we rely on the usage of a negative prior as described in Subsection~\ref{subsection:negprior}, if not stated otherwise. In our experiments, the negative prior is a standard normal with a shifted mean. For color images it is centered at $25 \cdot \mathbf{1}$, for grayscale images it is centered at $8 \cdot \mathbf{1}$. The magnitude of the shift is set based on a parameter sweep, which was evaluated using Fashion-MNIST and MNIST in the range of \{2, 4, 6, 8, 10\} for grayscale images, and using CIFAR-10 and SVHN in the range of \{5, 10, 15, 20, 25, 30\} for color images. After observing a clear trend, we have chosen the mode.

\input{table_color_summary}

\subsection{The effect of the noise model}

Examining the results for baseline VAE models (i.e., models without negative sampling) in Table~\ref{table:grayscale_summary} and Table~\ref{table:color_summary}, we can observe great variability in the OOD detection performance. 

The results suggest that the intriguing phenomenon in VAEs discussed by \cite{nalisnick2018deep} and \cite{choi2018waic} is highly dependent on modelling choices. In the case of grayscale images, when changing the noise model from Bernoulli to Gaussian --- and otherwise remaining in the same experimental setting as \cite{nalisnick2018deep}---, the issue of assigning higher likelihood estimates to OOD samples simply does not occur. However, one can observe that discrimination between inliers and OOD samples based on the KL divergence between approximate posterior and prior is hardly feasible, with below-$1/2$ AUC scores. Meanwhile, with a Bernoulli noise model --- also used in \cite{nalisnick2018deep}--- both the likelihood-estimates and the KL divergences fail to discriminate. The other results in the table where models are trained on MNIST confirm the asymmetric behaviour already described by \cite{nalisnick2018deep}, that is, switching the roles of the inlier and outlier dataset affects the presence of the phenomenon.
With color images, the corresponding rows of Table~\ref{table:color_summary} again show the importance of modelling choices.  When CIFAR-10 is the training set, the phenomenon persistently occurs with both Bernoulli and Gaussian noise models. When SVHN is the training set, one can observe again a great variability in the AUC scores.

\subsection{The effectiveness of the training method}

To demonstrate the effectiveness of training with negative samples, we present two different sets of experiments: first we incorporate negative samples from an auxiliary dataset, second we explore the use of adversarially generated negative samples from the trained model itself, and compare these to the standard training of VAE models.

\paragraph{Using auxiliary datasets}
The AUC scores in Table~\ref{table:grayscale_summary} show that using the auxiliary dataset as a source of negative samples in most cases resulted in models that are capable to distinguish nearly perfectly between inliers and OOD samples. This is also the case with color images, as experimental results in Table~\ref{table:color_summary} show.

\paragraph{Failure modes with auxiliary datasets} 
One can observe in Table~\ref{table:grayscale_summary} that --- despite the above mentioned improvements --- there are cases when using an auxiliary dataset fails to improve on the OOD separating capability. One example for this is when the inlier set is the EMNIST-Letters, the OOD test set is MNIST, and the used auxliary dataset is Fashion-MNIST, the results for this setup are in the last row of Table ~\ref{table:grayscale_summary}. 
We hypothesize, that this as an example of the case, when the auxiliary dataset (regarding its features) does not wedge in between the inlier and the outlier test set. One possible way of improvement in this regard is to utilize several auxiliary datasets to present a more diverse set of examples for possible OOD data in terms of features and semantic content.

\paragraph{Unsupervised method} 
In the case of the grayscale images, the last column in Table~\ref{table:grayscale_summary} shows the effectiveness of the fully unsupervised approach: regardless of whether using a Gaussian and a Bernoulli noise model, the trained models achieve higher AUC KL scores than the baseline in \emph{all permutations}. The method also shows better AUC BPD scores than the baseline in most of the cases where the baseline fails (i.e., baselines with below 0.6 AUC BPD scores). One can observe that when the train set is EMNIST-Letters and the OOD set is MNIST, the separation is still not achieved even with this method. 
The possible reason behind this is that the visual features of these two datasets are very close to each other and it is a hard task to switch the default relation between them. Note that when these two datasets switch roles, the likelihood estimates are correct. Table~\ref{table:color_summary} shows that in the case of color images, the unsupervised method also achieves notable discriminative performance
improving on the baseline.

\paragraph{Random noise and additive isotropic Gaussian noise does not help} 
We also investigated how the choice of negative samples influences the performance of the trained model. We conducted further experiments with the following negative samples: 1) Kuzushiji-MNIST (KMNIST) \cite{clanuwat2018deep} as another auxiliary dataset, 2) random noise, where we sample each pixel intensity from the uniform distribution on $[0,1]$ --- modeling a dataset with less structure, 3) with an additive isotropic Gaussian noise added to the inlier dataset.
The results in Table~\ref{table:negative_sample_sources} show that using the KMNIST also results in perfect separation of the inliers (Fashion-MNIST) and outliers (MNIST). The weak results with random noise as negative samples show the significance of the choice of negative samples. We also experimented with using as negative samples the training set itself with an additive isotropic Gaussian noise, a rather natural choice to provide near-manifold examples.
With an additive noise of $\sigma$=$0.25$, our results show weak discriminative power.

\input{table_negative_sources}

%% file: table_grayscale_summary.tex
\begin{table*}[!tb]
    \begin{center}
    \caption{Comparing the OOD discriminative power of baseline VAE models and VAE models with negative sampling on grayscale images.
    }
    \label{table:grayscale_summary}
    \begin{tabular}{ c  c  c  c  c  c  c }
      \toprule
      & \multirow{2}{*}{\textbf{Inlier}} & \multirow{2}{*}{\textbf{OOD}} & \multirow{2}{*}{\textbf{Noise model}}& \textbf{Baseline VAE} & \textbf{Negative:} & \textbf{Negative:} \\
      & & & & \textbf{(no negative)} & \textbf{auxiliary} & \textbf{adversarial}\\
      \midrule
      \parbox[t]{4mm}{\multirow{12}{*}{\rotatebox[origin=c]{90}{\textbf{AUC BPD}}}}
      &Fashion-MNIST & MNIST & Bernoulli & $0.46$ $(0.05)$ & $1.00$ $(0.00)$ & $0.70$ $(0.13)$ \\
      &Fashion-MNIST & MNIST& Gaussian & $0.98$ $(0.00)$ & $1.00$ $(0.00)$ & $0.80$ $(0.04)$ \\
      & Fashion-MNIST & EMNIST-Letters & Bernoulli & $0.61$ $(0.01)$ & $0.99$ $(0.00)$ & $0.78$ $(0.07)$ \\
      & Fashion-MNIST & EMNIST-Letters & Gaussian & $0.97$ $(0.00)$ & $1.00$ $(0.00)$ & $0.85$ $(0.04)$ \\
      \cmidrule{2-7}
      &MNIST & Fashion-MNIST & Bernoulli & $1.00$ $(0.00)$ & $1.00$ $(0.00)$ & $1.00$ $(0.00)$ \\
      &MNIST & Fashion-MNIST & Gaussian & $0.97$ $(0.00)$ & $1.00$ $(0.00)$ & $0.98$ $(0.01)$ \\
      &MNIST & EMNIST-Letters & Bernoulli & $0.99$ $(0.00)$ & $0.99$ $(0.00)$ & $0.98$ $(0.00)$ \\
      &MNIST & EMNIST-Letters & Gaussian & $0.78$ $(0.14)$ & $0.93$ $(0.08)$ & $0.79$ $(0.04)$ \\
      \cmidrule{2-7}
      &EMNIST-Letters & Fashion-MNIST & Bernoulli & $0.98$ $(0.00)$ & $0.98$ $(0.00)$ & $0.99$ $(0.00)$ \\
      &EMNIST-Letters & Fashion-MNIST & Gaussian & $0.80$ $(0.07)$ & $0.76$ $(0.08)$ & $0.93$ $(0.04)$ \\
      &EMNIST-Letters&  MNIST & Bernoulli & $0.58$ $(0.02)$ & $0.58$ $(0.02)$ & $0.73$ $(0.07)$ \\
      &EMNIST-Letters & MNIST & Gaussian & $0.67$ $(0.17)$ & $0.58$ $(0.20)$ & $0.65$ $(0.04)$ \\
      \midrule
      \parbox[t]{4mm}{\multirow{12}{*}{\rotatebox[origin=c]{90}{\textbf{AUC KL}}}}
      &Fashion-MNIST & MNIST & Bernoulli & $0.61$ $(0.09)$ & $1.00$ $(0.00)$ & $0.88$ $(0.07)$ \\
      &Fashion-MNIST & MNIST & Gaussian & $0.26$ $(0.03)$ & $1.00$ $(0.00)$ & $0.74$ $(0.05)$ \\
      &Fashion-MNIST & EMNIST-Letters & Bernoulli & $0.68$ $(0.07)$ & $1.00$ $(0.00)$ & $0.84$ $(0.04)$ \\
      &Fashion-MNIST & EMNIST-Letters & Gaussian & $0.38$ $(0.04)$ & $0.99$ $(0.00)$ & $0.79$ $(0.05)$ \\
      \cmidrule{2-7}
      &MNIST & Fashion-MNIST & Bernoulli & $0.73$ $(0.14)$ & $1.00$ $(0.00)$ & $0.94$ $(0.10)$ \\
      &MNIST & Fashion-MNIST & Gaussian & $0.71$ $(0.04)$ & $1.00$ $(0.00)$ & $0.98$ $(0.01)$ \\
      &MNIST & EMNIST-Letters & Bernoulli & $0.64$ $(0.03)$ & $0.76$ $(0.03)$ & $0.89$ $(0.02)$ \\
      &MNIST & EMNIST-Letters & Gaussian & $0.54$ $(0.07)$ & $0.75$ $(0.08)$ & $0.74$ $(0.04)$ \\
      \cmidrule{2-7}
      &EMNIST-Letters & Fashion-MNIST & Bernoulli & $0.66$ $(0.14)$ & $0.54$ $(0.09)$ & $0.98$ $(0.00)$ \\
      &EMNIST-Letters & Fashion-MNIST & Gaussian & $0.54$ $(0.10)$ & $0.49$ $(0.23)$ & $0.91$ $(0.05)$ \\
      &EMNIST-Letters & MNIST & Bernoulli & $0.37$ $(0.05)$ & $0.45$ $(0.03)$ & $0.75$ $(0.06)$ \\
      &EMNIST-Letters & MNIST & Gaussian & $0.36$ $(0.08)$ & $0.43$ $(0.10)$ & $0.64$ $(0.04)$ \\
      \bottomrule
    \end{tabular}
    \end{center}
\end{table*}

%% file: table_color_summary.tex
\begin{table*}[!tb]
    \begin{center}
    \caption{Comparing the OOD discriminative power of baseline VAE models and VAE models with negative sampling on color images.}
    \label{table:color_summary}
    \begin{tabular}{ c  c  c  c  c  c  c }
      \toprule
      & \multirow{2}{*}{\textbf{Inlier}} & \multirow{2}{*}{\textbf{OOD}} & \multirow{2}{*}{\textbf{Noise model}}& \textbf{Baseline VAE} & \textbf{Negative:} & \textbf{Negative:} \\
      & & & & \textbf{(no negative)} & \textbf{auxiliary} & \textbf{adversarial}\\
      \midrule
      \parbox[t]{4mm}{\multirow{4}{*}{\rotatebox[origin=c]{90}{\textbf{AUC BPD}}}}
      &CIFAR-10 & SVHN & Bernoulli &  $0.59$ $(0.00)$ & $0.90$ $(0.05)$ &$0.81$ $(0.04)$ \\
      &CIFAR-10 & SVHN & Gaussian & $0.25$ $(0.02)$ & $0.93$ $(0.01)$ & $0.84$ $(0.03)$ \\
      \cmidrule{2-7}
      &SVHN  & CIFAR-10 & Bernoulli &  $0.51$ $(0.00)$ & $1.00$ $(0.00)$& $0.70$ $(0.03)$\\
      &SVHN & CIFAR-10 & Gaussian & $0.92$ $(0.00)$ &
      $1.00$ $(0.00)$ & $0.75$ $(0.11)$\\
      \midrule
      \parbox[t]{4mm}{\multirow{5}{*}{\rotatebox[origin=c]{90}{\textbf{AUC KL}}}}
      &CIFAR-10 & SVHN & Bernoulli & $0.29$ $(0.00)$ & $0.90$ $(0.06)$ & $0.81$ $(0.04)$  \\
      &CIFAR-10 & SVHN & Gaussian &$0.25$ $(0.01)$ & $0.93$ $(0.01)$ &  $0.84$ $(0.03)$ \\
      \cmidrule{2-7}
      &SVHN  & CIFAR-10 & Bernoulli &  $0.87$ $(0.00)$ &
      $1.00$ $(0.00)$ & $0.70$ $(0.03)$\\
      &SVHN & CIFAR-10 & Gaussian &$0.74$ $(0.01)$ &	
      $1.00$ $(0.00)$  & $0.74$ $(0.11)$ \\
      \bottomrule
    \end{tabular}
    \end{center}
\end{table*}

%% file: table_negative_sources.tex
\begin{table*}[!tb]
    \caption{Comparing baseline VAE model and training with negative samples from different sources.}
    \label{table:negative_sample_sources}
    \begin{center}
    \begin{tabular}{ l c c  c c c c c c c }
      \toprule
      \textbf{Inlier} & \textbf{OOD} & & \textbf{Baseline VAE} & \multicolumn{3}{c}{\textbf{Auxiliary data as negative}} & \textbf{Negative:} \\
      \textbf{Fashion-MNIST}  &\textbf{MNIST} & & \textbf{(no negative)}& \textbf{Random} & \textbf{Inlier + Gaussian noise} & \textbf{KMNIST} & \textbf{adversarial} \\
      \midrule 
      \textbf{AUC BPD} & & &$0.46$ $(0.05)$ & $0.47$ $(0.05)$ & $0.44$ $(0.01)$ & $1.00$ $(0.00)$ &$0.70$ $(0.13)$  \\ 
      \textbf{AUC KL} & & & $0.61$ $(0.09)$ & $0.56$ $(0.08)$ & $0.70$ $(0.09)$ & $1.00$ $(0.00)$ &$0.88$ $(0.07)$ \\
      \bottomrule
    \end{tabular}
    \end{center}
\end{table*}

%% file: 5_related_work.tex
\section{Related Work}

Our investigations are mostly inspired by and related to recent work on the evaluation of generative models on OOD data \cite{ nalisnick2018deep, choi2018waic, hendrycks2018deep}. These works reported firstly that despite expectations, generative models --- including, but not limited to VAEs --- consistently fail at distinguishing OOD data from the training data, yielding higher likelihood estimates on unseen OOD samples.
Reference \cite{nalisnick2018deep} examines the phenomenon in detail, focusing on finding the cause of it by analyzing flow-based models that allow exact likelihood calculation. In \cite{choi2018waic} also notice the above-mentioned phenomenon, while they address the task of OOD sample detection with Generative Ensembles. They decrease the weight of the KL divergence term in the ELBO to encourage a higher distortion penalty during training, resulting in a better performing model. This observation also confirms the importance of the noise model and the balance between the KL term and the reconstruction term.
The ominous observation is presented also by \cite{hendrycks2018deep}, but they concentrate on improving the OOD data detection with Outlier Exposure. Their work demonstrates that using samples from an auxiliary data set as OOD examples, significantly improves on the performance of existing OOD detection models on image and text data. However, they do not investigate the VAE model in detail, and their general setup always requires an auxiliary dataset. Our work also sheds light on an issue with this approach: one should choose the auxiliary datasets carefully to obtain robust OOD detection. This paper is updated version of our unpublished preprint \cite{csiszarik2019negative}. 

More recent work, \cite{xiao2020likelihood} also emphasises that previous OOD detection scores do not always improve the performance of VAEs and introduces as new score called the Likelihood Regret.Their score requires the optimization of the encoder parameters for each individual test sample separately, which introduces considerable computational overhead.

Within the context of uncertainty estimation, \cite{lee2017training} demonstrate that adversarially generated samples improve the confidence of classifiers in their correct predictions. They train a classifier simultaneously with a GAN and require it to have lower confidence on GAN samples. For each class distribution, they tune the classifier and GAN using samples from that OOD dataset. Their method of using generated samples of GANs is closest to our approach of using generated data as negative samples, but \cite{lee2017training} work within a classification setting.

Complementary to our work, \cite{daniel2019deep} also introduces the training VAEs with negative examples, but they focus their analysis on semi-supervised setting. We instead aim at improving the misleading likelihood estimates of VAEs, especially in fully unsupervised manner with adversarial training scheme. 
The idea of introducing an additional prior for outliers also appears in \cite{ran2022detecting}. Their  
objective is the sum of ELBOs for inliers and outliers, with the additional reconstruction term of OOD samples compared to our $\mathcal{L}_{np}$ loss from Subsection~\ref{subsection:negprior}, though learning to reconstruct outliers may not always be beneficial. To obtain near-manifold OOD examples, they add Gaussian noise to in-distribution data --- interestingly, in our setup and experiments summarized in Table~\ref{table:negative_sample_sources}, this did not aid discrimination.

%% file: 6_conclusion.tex
\section{Conclusion}

In this work, we studied Variational Autoencoder models and investigated to what extent the latent representations of data points or the likelihood estimates given by the model can be used to identify out-of-distribution samples. 
We presented empirical evidence that utilizing negative samples either from an auxiliary dataset or from an adversarial training scheme significantly and consistently improves the discriminative power of VAE models regarding out-of-distribution samples.